\documentclass[lettersize,journal]{IEEEtran}
\usepackage{cite}
\usepackage{amsmath,amssymb,amsfonts}
\usepackage{algorithmic}
\usepackage{graphicx}
\usepackage{multirow}
\usepackage{caption}
\usepackage{booktabs}
\usepackage{textcomp}
\usepackage{xcolor}
\usepackage{amssymb}
\usepackage[switch]{lineno}
\usepackage[flushleft]{threeparttable}
\def\BibTeX{{\rm B\kern-.05em{\sc i\kern-.025em b}\kern-.08em
    T\kern-.1667em\lower.7ex\hbox{E}\kern-.125emX}}

\DeclareCaptionFont{normal}{\normalfont}
\usepackage[
  colorlinks=false,                   
  citebordercolor={0 1 0},                
  pdfborder={0 0 1.5},
]{hyperref}

\begin{document}

\newcommand{\cmt}[1]{{\color{red}#1}}
\newcommand{\method}{HierRelTriple}

\title{\method: Guiding Indoor Layout Generation with Hierarchical Relationship Triplet Losses \\
}

\author{%
  Kaifan~Sun,
  Bingchen~Yang,
  Peter~Wonka,
  Jun~Xiao,
  Haiyong~Jiang
}

\twocolumn[{
\renewcommand\twocolumn[1][]{#1}
\maketitle
\begin{center}
    \captionsetup{type=figure}
    \includegraphics[width=\textwidth]{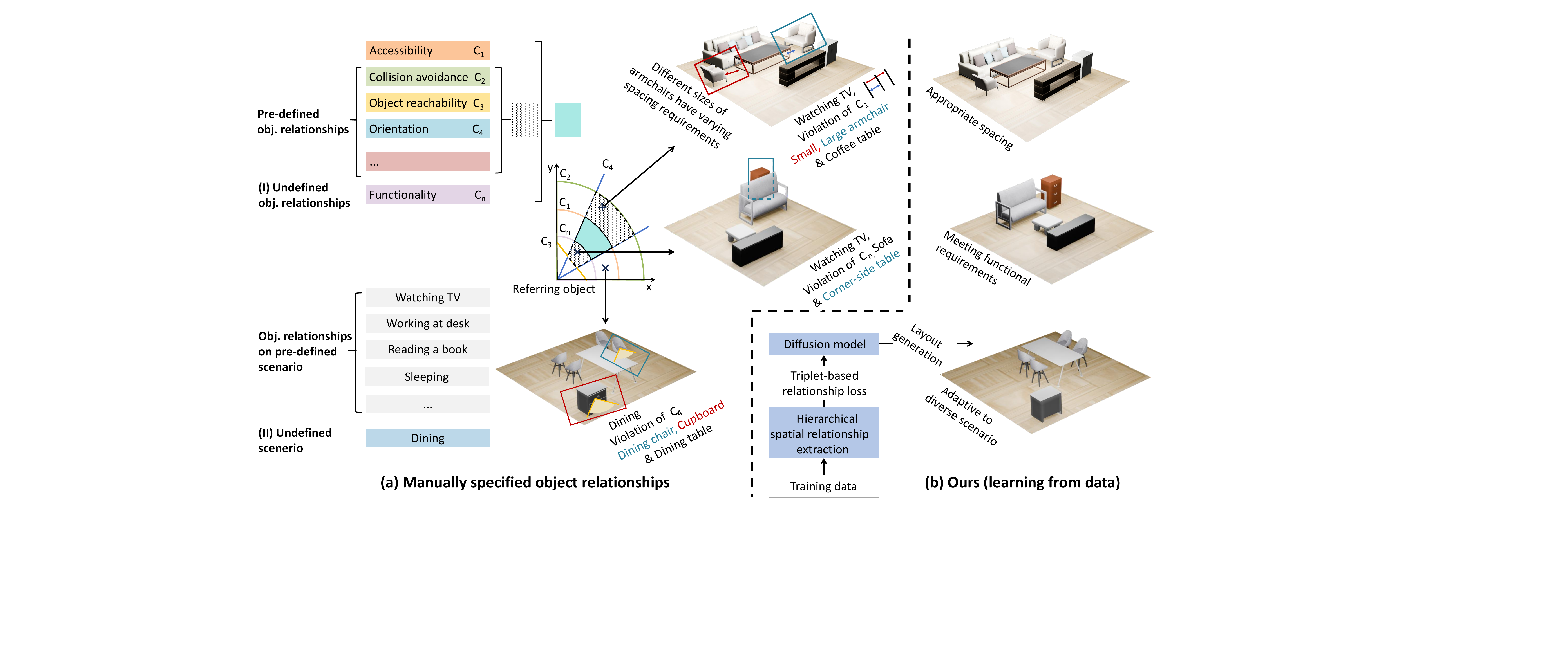}
    \captionof{figure}{Comparison of indoor layout synthesis methods: (a) Manually specified relationships tend to be incomplete, resulting in unrealistic layout generation. (b) Our data-driven relationship learning derives important spatial relationships automatically from the dataset, enabling the creation of well-organized layouts.}
    \label{fig:teaser}
\end{center}
}]
\begingroup
\renewcommand\thefootnote{} 
\footnotetext{%
\hspace{-3mm}Kaifan Sun, Bingchen Yang, Jun Xiao, and Haiyong Jiang are with the
  University of Chinese Academy of Sciences (UCAS), Beijing~100049,~China
  (e-mail: \{sunkaifan22, yangbingchen211\}@mails.ucas.ac.cn, \{xiaojun, 
haiyong.jiang\}@ucas.ac.cn).
  Peter Wonka is with the King Abdullah University of Science and Technology
  (KAUST), Thuwal 23955, Saudi Arabia
  (e-mail: pwonka@gmail.com). Haiyong is the Project Lead.}
\footnotetext{\hspace{-3mm}Kaifan Sun and Bingchen Yang contributed equally.}
\footnotetext{\hspace{-3mm}Haiyong Jiang and Jun Xiao are the corresponding authors.}
\endgroup
\nolinenumbers
\begin{abstract}
We present a hierarchical triplet-based indoor relationship learning method, coined \method{}, with a focus on spatial relationship learning.
Existing approaches often depend on manually defined spatial rules or simplified pairwise representations, which fail to capture complex, multi-object relationships found in real scenarios and lead to overcrowded or physically implausible arrangements. We introduce \method{}, a hierarchical relational triplets modeling framework that first partitions functional regions and then automatically extracts three levels of spatial relationships: object-to-region (O2R), object-to-object (O2O), and corner-to-corner (C2C). By representing these relationships as geometric triplets and employing approaches based on Delaunay Triangulation to establish spatial priors, we derive IoU loss between denoised and ground truth triplets and integrate them seamlessly into the diffusion denoising process. 
The introduction of the joint formulation of inter-object distances, angular orientations, and spatial relationships enhances the physical realism of the generated scenes.
Extensive experiments on unconditional layout synthesis, floorplan-conditioned layout generation, and scene rearrangement demonstrate that \method{} improves spatial-relation metrics by over 15\% and substantially reduces collisions and boundary violations compared to state-of-the-art methods.
\end{abstract}



\section{introduction}

Automatically generating indoor layouts is critical for a range of applications, including virtual and augmented reality (VR/AR)~\cite{smelik2014survey}, smart home systems~\cite{ma2022smart}, robotics~\cite{fernandez2022robot}, and interior designs~\cite{fisher2012example,zhang2022sceneviewer,fu2023plannet}. 
The rise of generative models~\cite{wang2021sceneformer, paschalidou2021atiss, wei2023lego, maillard2024debara} has unlocked new possibilities, significantly advancing both quality and efficiency of indoor layout generation.
Despite the progress, generating layouts with practical usability and proper spacing relationships among objects and floorplan boundaries remains an open challenge. Achieving this goal requires a compilation of physical constraints, ergonomic rules, walking spacing, co-occurring object placements, and so on.

Recent approaches~\cite{tang2024diffuscene, yang2024physcene, sun2024haisor, leimer2022layoutenhancer, hong2024human} derive primary object relationships from expert knowledge to aid layout generation, such as collision avoidance~\cite{tang2024diffuscene, yang2024physcene}, object reachability~\cite{yang2024physcene, leimer2022layoutenhancer}, and human behavior guidance~\cite{sun2024haisor, hong2024human}.
These relationships are constructed and enforced as additional regularizations~\cite{tang2024diffuscene, leimer2022layoutenhancer} and classifier guidance on a generative model~\cite{yang2024physcene}.
However, most methods focus on specific relationships based on predefined rules and parameters and usually do not consider the proper spacing between objects and an object to a functional region. 
For example, PhyScene~\cite{yang2024physcene} emphasizes collision avoidance and reachability, but there is no regularization to maintain object spacing between functionally related objects. Forest2Seq~\cite{sun2024forest2seq} establishes hierarchical object relationships but provides insufficient modeling of the relative relationships among sibling objects. As shown in Fig.~\ref{fig:teaser} (a), armchairs of different sizes impose distinct spacing requirements. A sofa can be positioned too close to the corner-side table while being far from the coffee table and TV stand. 
Another work, LayoutEnhancer~\cite{leimer2022layoutenhancer}, lacks spacing relations to floorplan boundaries and functional regions, leading to boundary violations between objects and the wall.
Moreover, category-wise relationship modeling, e.g., LayoutEnhancer, requires tedious manual efforts to determine affected objects by each relationship type and is not scalable.

In light of these problems, we directly learn object spacings from the training data as illustrated in Fig.~\ref{fig:teaser} (b).
The spacing between nearby objects within a functional region is usually important to ensure functional pathways and physical interactivity. In addition, the spacing between an object and the functional region bounds can avoid boundary violations and unreasonable object positions.
To facilitate automatic relationship extraction, we observe that the three-level hierarchy, i.e., furniture-functional region-layout, can partition objects within a layout and that the geometric properties of the Delaunay Triangulation (DT) facilitate grouping proximate object triplets as a DT triangle, and emphasizing objects with a larger shape size with DT weights during relationship learning. 

From another perspective, previous works~\cite{tang2024diffuscene, yang2024physcene, leimer2022layoutenhancer} mainly model pairwise relationships between two objects, making it difficult to account for high-order object interactions and important indoor design principles.
A typical example is the kitchen work triangle~\cite{lange2012woman} that addresses the efficiency between major work centers, including cooking, sink, and refrigerator.
Similar examples can be found in a living room, e.g., the triangular layout among sofa, armchairs, and TVs. 
Therefore, introducing high-order triplet-based relationships can consider interactions among three objects and constrain relative positions better.

We propose a novel framework, \method{}, to render the two above-mentioned observations to integrate spatial relationships into a generative model.
The overall pipeline of \method{} is shown in Fig.~\ref{fig:pipeline} and is based on DiffuScene~\cite{dhariwal2021diffusion}.
First, we automatically extract important relationships from an input layout hierarchically for training.
The relationship extraction begins by grouping nearby objects into functional regions. 
We categorize spatial relationships into three types, including relative placement of an object with respect to its belonging functional region, the spacing between the centroids of two objects within a functional region, and nearby relationships between corners of any two objects. 
We further subdivide these relationships into a set of triplets using DT. 
Thereafter, we present a unified framework that enforces triplet-based relationships with an IoU loss. 
The diffusion process is optimized using both the diffusion loss and the proposed losses for relationship learning. 
The source code and trained models will be released. 
Our main contributions are as follows.
\begin{enumerate}
   \item A method to automatically extract important object-to-region, object-to-object, and corner-to-corner relationships as triplets from an input floorplan. 
    \item An IoU-based loss for the learning of triplet relationships during the training of a diffusion model.
    \item Superior performance on established scene synthesis metrics and significant improvements (at least $15\%$) over existing methods on newly introduced spatial relationship metrics for three tasks.
\end{enumerate}


\section{related work}\label{sec:related}

In this section, we first review generative indoor layout modelling methods, including approaches based on VAEs, GANs, autoregressive models, and diffusion. We then discuss graph-based relationship learning methods that capture high-level spatial and semantic dependencies. Finally, we examine optimisation-based strategies for integrating fine-grained functional and geometric constraints into layout generation.

\subsection{Generative Indoor Layout Modeling}
The success of generative models in image generations~\cite{ramesh2021zero,rombach2022high} has encouraged likewise explorations in indoor layout modelling, including variational autoencoder (VAE)-based methods~\cite{luo2020end, gao2023scenehgn, zhai2023commonscenes}, GAN-based methods~\cite{yang2021indoor, koh2023simple, bahmani2023cc3d, xu2023discoscene}, autoregressive models~\cite{ritchie2019fast, wang2019planit, para2021generative, wang2021sceneformer, nie2023learning,paschalidou2021atiss,lewis2019bart}, and diffusion models~\cite{wei2023lego, tang2024diffuscene, lin2024instructscene, yang2024physcene, zhai2024echoscene, Yang2025MMGDreamerMG}.
Another interesting trend~\cite{fu2024anyhome, cCelen2024IDesignPL, feng2023layoutgpt, yang2024llplace, ding2023task, tam2024smc, wang2024chat2layout} is to leverage autoregressive Large Language Models (LLMs) as visual scene planners, generating layouts as structured text programs. 
Some interesting works~\cite{wei2023lego,yang2024physcene, zhai2024echoscene,tang2024diffuscene} explore diffusion models for layout rearrangement and conditional and unconditional indoor layout modelling. 
While these diverse generative approaches have demonstrated significant capabilities in synthesising plausible furniture arrangements, their primary focus often lies in generating sets of objects with plausible categorical and geometric attributes (e.g., type, size, position) within a scene. However, explicit and structured modelling of the spatial relationships between objects remains a less emphasised or implicitly learnt aspect in many existing frameworks. Compared to the methods above, this work focuses on modelling spatial relationships with inter-object functional requirements.
\begin{figure*}[htbp]
    \centering
    \includegraphics[width=0.98\textwidth]{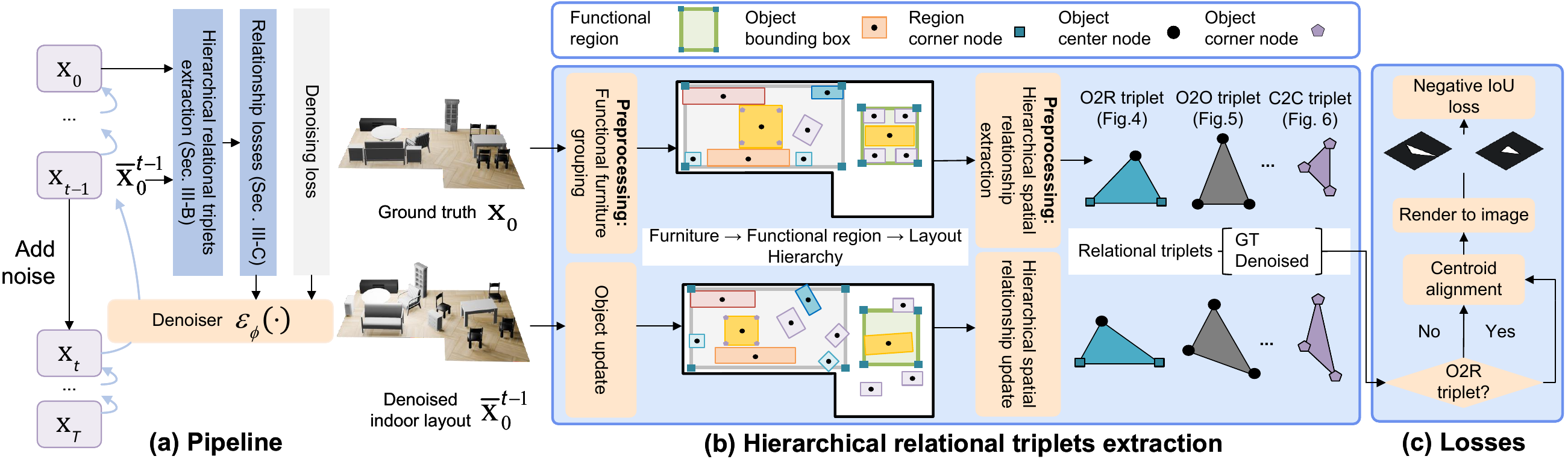}
        \caption{Overview of \method{}. (a) Spatial relationships are automatically extracted and modeled, then integrated into diffusion-based model training through a rendering loss. (b) A hierarchical approach is applied to identify important spatial relationships. (c) Evaluation of these losses is performed through differentiable rendering.
      }
    \label{fig:pipeline}  
\end{figure*}
\subsection{Graph-based Relationship Learning for Indoor Layouts}
Plausible indoor layouts are usually bound with complex layout relationships~\cite{de2001time}, such as co-occurrence, functionality, feasible placement, and layout balance.
One line of work abstracts layout relationships as a graph for graph-conditioned layout modelling.
The relationship graph can be manually specified~\cite{luo2020end, dhamo2021graph, xu2023scene,  zhai2023commonscenes, zhai2024echoscene, wei2024compositional} or generated~\cite{wang2019planit, li2019grains, para2021generative, gao2023scenehgn, chattopadhyay2023learning,lin2024instructscene, zhang2024sceneexplorer,bai2025freescene}, with nodes representing furniture objects annotated with attributes and edges encoding targeted spatial and semantic relationships (e.g., orientation, functional grouping). However, while effectively propagating high-level relational semantics and topological structures, these methods often lack explicit mechanisms to satisfy fine-grained, geometric constraints inherent in spatial relationships. 
Generated geometries (e.g., positions, orientations) may statistically approximate relational priors learnt from data, but critical functional constraints (e.g., maintaining minimal clearance distances or strictly avoiding object interpenetration) are missed. 
Compared to these graph-conditioned approaches, our work learns relationships with hierarchically constructed relationship losses and the diffusion process.
\subsection{Optimization-based Indoor Relationship Integration}

Spatial relationships are crucial for physically and functionally plausible indoor layouts. The core challenge lies in effectively integrating these relationships into the optimization process.
A common strategy involves optimizing a density function using Markov Chain Monte Carlo (MCMC)~\cite{merrell2011interactive, yeh2012synthesizing, yu2011make} or position-based dynamics~\cite{weiss2018fast,zhang2021fast,zhang2023scenedirector}, which enable satisfying fundamental constraints like collision avoidance and containment within boundaries.

Another scheme incorporates additional relationships by deep reinforcement learning~\cite{sun2024haisor} and additional conditional guidance on deep generative models. PhyScene~\cite{yang2024physcene} and SHADE~\cite{Hong2024HumanAware3S}  adopt the classifier-guidance diffusion model~\cite{dhariwal2021diffusion} to optimize indoor layouts with collision avoidance and object reachability. Ergonomic rules and human interaction constraints have been formulated as differentiable objective functions that are optimized directly within the training loop of generative models~\cite{leimer2022layoutenhancer}, targeting functional aspects like human movement paths and interaction.

Compared to these relation-wise formulations, which often require specialized handling for each constraint type (e.g., separate collision functions ~\cite{yang2024physcene,tang2024diffuscene}, reachability metrics~\cite{leimer2022layoutenhancer, yang2024physcene}, or path planning objectives~\cite{sun2024haisor}), our approach encodes diverse object relationships using a unified triple representation, which is easily integrated with rendering-based loss functions for efficient joint optimization.




\section{method}
Our goal is to learn a generative model that can generate indoor layouts with plausible object relationships. 
The overall pipeline is illustrated in Fig.~\ref{fig:pipeline}. 
We start by introducing the layout representation and the layout diffusion model in Sec.~\ref{subsec:prelim}. 
To further learn indoor spatial relationships, we proceed to build three categories of relational triplets using the training data (see Sec.~\ref{subsec:relation}). These include mutual relationships linking objects to their respective functional regions, relationships between distinct objects, and relationships between object corners.
We enforce these relationships using differentiable rendering-based losses and train the diffusion model by combining it with the diffusion loss (see Sec.~\ref{subsec:loss}). 
During inference, our method directly generates an indoor layout via the diffusion model without reliance on the spatial relationship construction. 

\subsection{The Layout Diffusion Model} \label{subsec:prelim}

We encode an indoor layout as \( \mathbf{x} = \{o_1, o_2, \dots, o_N\} \) with $N$ objects, where $N$ varies for different layouts.
Each object \( o_i \) has a set of parameters \( o_i = \{c_i, b_i, f_i\} \), where \( c_i \in \mathbb{R}^C \) indicates its category within a set of \( C \) categories. $b_i = (s_i, p_i, r_i)$ is the 3D bounding box of the object $o_i$, defined by its size \( s_i \in \mathbb{R}^3 \), the coordinate of centroid \( p_i \in \mathbb{R}^3 \), and the ground plane orientation ~\( r_i = (\cos\theta_i, \sin\theta_i) \in \mathbb{R}^2 \).  
\( f_i \in \mathbb{R}^{32} \) denotes a 3D object feature encoding the shape of the object and can be used for furniture retrieval.

Our method is built based on DiffuScene~\cite{tang2024diffuscene} and can easily incorporate other indoor layout generative models, e.g., autoregressive model-based ATISS.
Starting with a clean indoor layout \(\mathbf{x}_0 = \mathbf{x}\), we gradually add Gaussian noise through a forward diffusion process, transforming \(\mathbf{x}_0\) into noisy data \(\mathbf{x}_T\). This forward process consists of a series of transitions $q(\mathbf{x}_{t} \mid \mathbf{x}_{t-1})$ as follows:
\begin{equation}
q(\mathbf{x}_t \mid \mathbf{x}_{t-1}) = \mathcal{N}(\mathbf{x}_t; \sqrt{\alpha_t} \mathbf{x}_{t-1},  (1 - \alpha_t)\mathbf{I})
\end{equation}
$\beta_t$ controls the noise variance at each time step.
During the generative process, we gradually denoise noisy layout \(\mathbf{x}_T \sim \mathcal{N}(0, \mathbf{I})\) with the reverse process $p(\mathbf{x}_{t-1} \mid \mathbf{x}_t)$. 
The reverse process is optimized by maximizing the log-likelihood of $p(\mathbf{x}_0)$:
\begin{equation}
\mathcal{L}_{\text{diff}} = \mathbb{E}_{\mathbf{x}_0, \epsilon, t} \left[ \lVert \epsilon - \epsilon_\phi(\mathbf{x}_t, t) \rVert_2^2 \right],
\label{eq:diffuse} 
\end{equation}
where $\epsilon$ denotes the added Gaussian noise in each step, and $\epsilon_\phi(\mathbf{x}_t, t)$ is a denoising network. 
The denoising network $\epsilon_\phi(\mathbf{x}_t, t)$ adopts a UNet-1D module with self-attention \cite{vaswani2017attention}. 

During training, although the denoising model predicts the noise $\epsilon_\phi(\mathbf{x}_t, t)$ at time step $t$, we can deterministically (without iterative sampling) recover the denoised layout $\bar{\mathbf{x}}^t_0$ (an estimate of the ground-truth $\mathbf{x}_0$) from Eq.~\ref{Eq:denoised_layout} as,
\begin{equation}
    \bar{\mathbf{x}}_0^t = \frac{1}{\sqrt{\bar{\alpha_t}}}\big(\mathbf{x}_t - \sqrt{1-\bar{\alpha_t}}\epsilon_\phi(\mathbf{x}_t, t) \big)
    \label{Eq:denoised_layout}
\end{equation}
where $\bar{\alpha}_t = \prod_{i=1}^t\alpha_i$. This denoised layout will enable us to compute the relationship losses in Sec.~\ref{subsec:loss}.

\subsection{Hierarchical Relational Triplets Extraction}\label{subsec:relation}

Indoor spatial relationships, e.g., walking space maintenance, collision avoidance, and co-occurrence between nearby objects, are usually organized hierarchically, following the object-functional region-layout hierarchy of indoor scenes as illustrated in Fig.~\ref{fig:pipeline} (b).
For example, a sofa and a corner table in the living room need to maintain appropriate spacing to accommodate proper interactions between the table and the human on the sofa, while subdivided regions within the layout restrict the spatial boundaries of objects, such as a dining table within the dining region.
To extract hierarchical spatial relationships, we first extract functional regions to construct a three-level indoor hierarchy and then divide spatial relationships into three types. 
This process can be done as a one-time preprocessing step for each input layout and is only executed for the training stage.

\begin{figure}[htbp]
    \centering
    \includegraphics[width=0.95\linewidth]{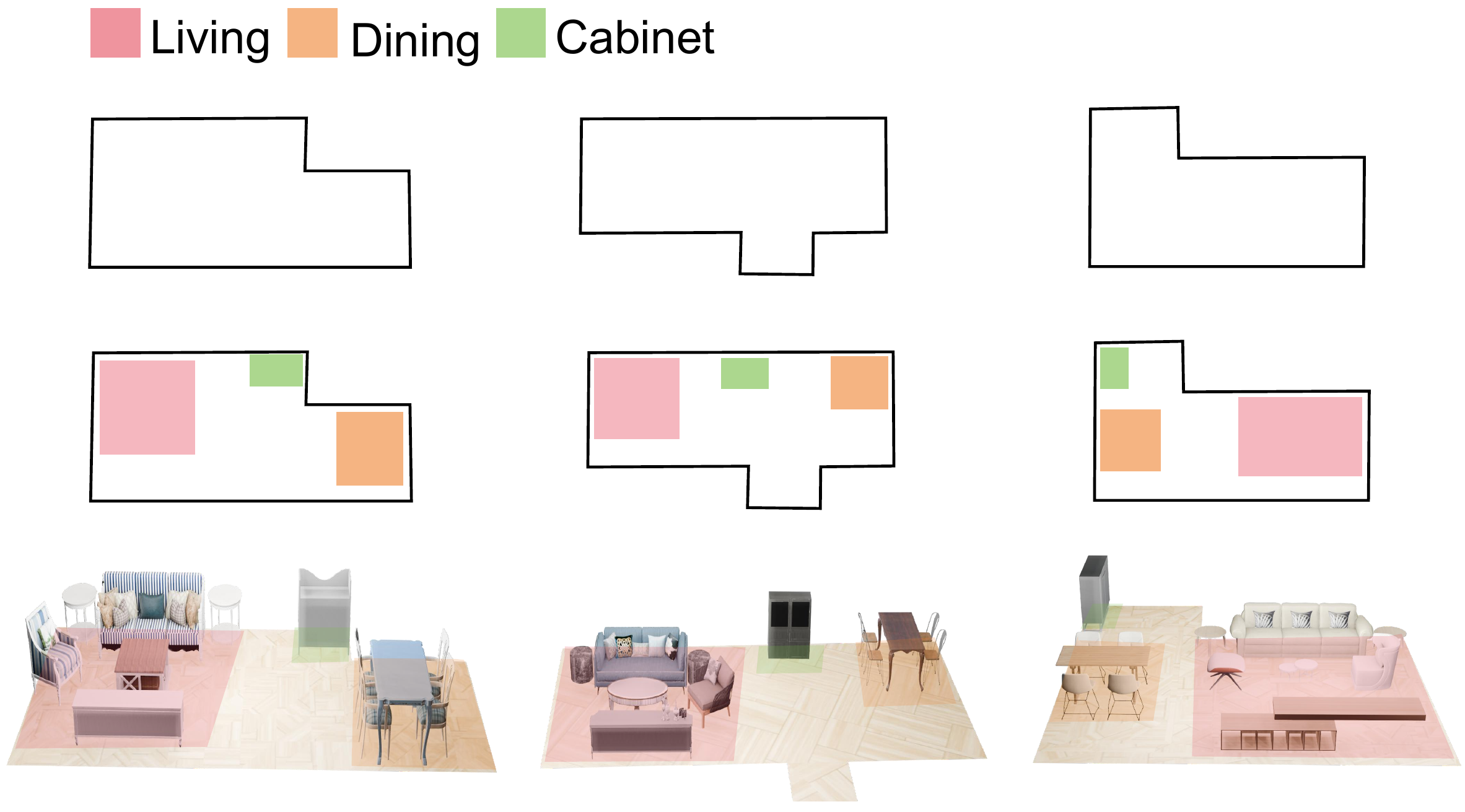}
        \caption{Three functional regions (Living, Dining, and Cabinet) are identified within the scene through color-based delineation.}
    \label{fig:region}  
\end{figure}

To construct functional regions, we project the indoor layout onto the ground plane and represent objects as vertices. 
We group objects as functional regions using clustering and encode functional regions as rectangles.
Following SceneHGN~\cite{gao2023scenehgn}, we utilize DBSCAN~\cite{ester1996density} to delineate object clusters based on the overlapping ratio and the distance criteria between any two objects $o_i, o_j$:
\begin{equation}
\label{eq:DBSCAN}
m_{ij} = d_{ij} + \lambda_{\text{giou}} \cdot (1 - \text{GIoU}(b_i, b_j)),
\end{equation}
\noindent where $\text{GIoU}(\cdot) \in [-1, 1]$~\cite{rezatofighi2019generalized} takes into account the influence of bounding boxes $b_i, b_j$ of two objects and $d_{ij}=||p_i - p_j||^2_{xy}$ is the Euclidean distance between the centroids of the two objects on the ground plane. 
We define the functional regions as the bounding box of each cluster. The vertex coordinates of each region are derived from the maximum-minimum operator applied on the centroid coordinates of all objects within the cluster. 
Fig.~\ref{fig:region} shows some constructed functional regions.

After obtaining the functional regions, we then extract hierarchical spatial relationships. 
Previous arts~
\cite{yang2024physcene,leimer2022layoutenhancer,sun2024haisor} optimize indoor relationships according to different formulations of semantics and functionality, such as collision avoidance, walking space, and functional relevance.
These formulations predominantly boil down to optimizing spatial distances between objects. 
Our approach diverges by optimizing spatial relationships according to their relevance within the indoor hierarchy. 
For example, collision avoidance mainly constrains the relative distance between object corners, while functionally related objects usually follow specific spatial patterns between object positions.
We build three types of spatial relationships: object-to-functional region (O2R) relationships that encode the spatial relationship between individual objects and layout regions, object-to-object (O2O) relationships that encode relative distances among the centroids of objects, and corner-to-corner (C2C) relationships that restrict distances between two objects' corners.

\begin{figure}[htbp]
    \centering
    \includegraphics[width=0.98\linewidth]{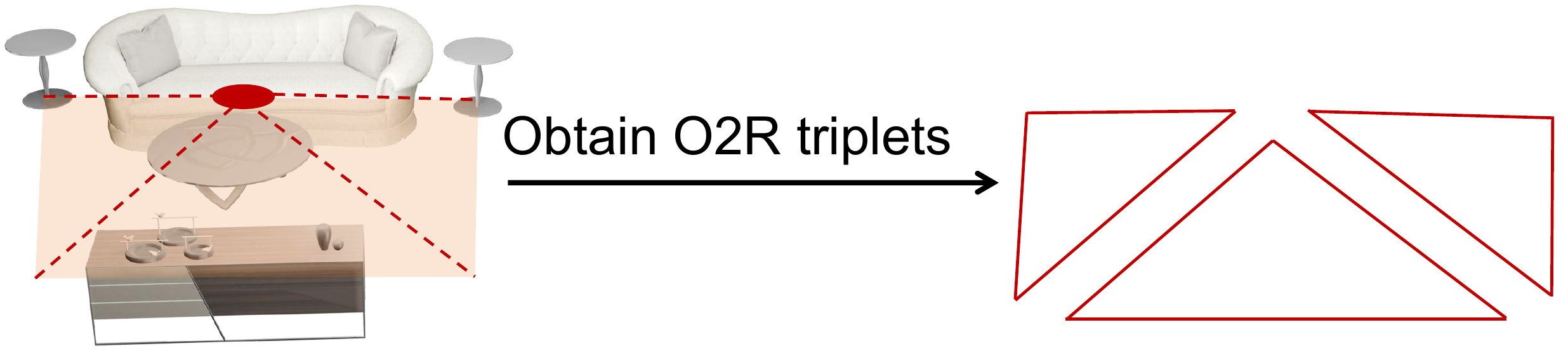}
        \caption{An example illustrating the construction of an O2R triplet.}
    \label{fig:O2R}  
\end{figure}

\noindent\textbf{O2R Triplets.} 
O2R relationships model the relative spatial distribution of a specific object concerning its belonging functional region.  
We treat the four corners of a functional region, determined by its bounding box, as virtual nodes. 
The furniture object with the largest size, which is usually the dominant object of a functional region, is then connected to these four virtual nodes of its associated functional region, forming four triangles. 
These triangles, each composed of one real furniture object and two virtual corner points, constitute the O2R triplets. 
We filter out elongated triangles containing any interior angle exceeding 160$^\circ$.
A sample illustrating the construction of O2R triples is shown in Fig.~\ref{fig:O2R}.
These triplets serve as a fundamental mechanism to explicitly model the critical spatial relationships that must be maintained between furniture items and the boundaries of their designated functional regions within the room. This modeling encompasses essential constraints such as ensuring furniture avoids breaching room or region boundaries, thereby maintaining safe and practical clearances. It also includes guiding the placement of key, dominant objects, exemplified by a bed in a bedroom or a sofa in a living room, into the central area of their respective regions to establish visual balance and functional prominence. 

\begin{figure}[htbp]
    \centering
    \includegraphics[width=0.46\textwidth]{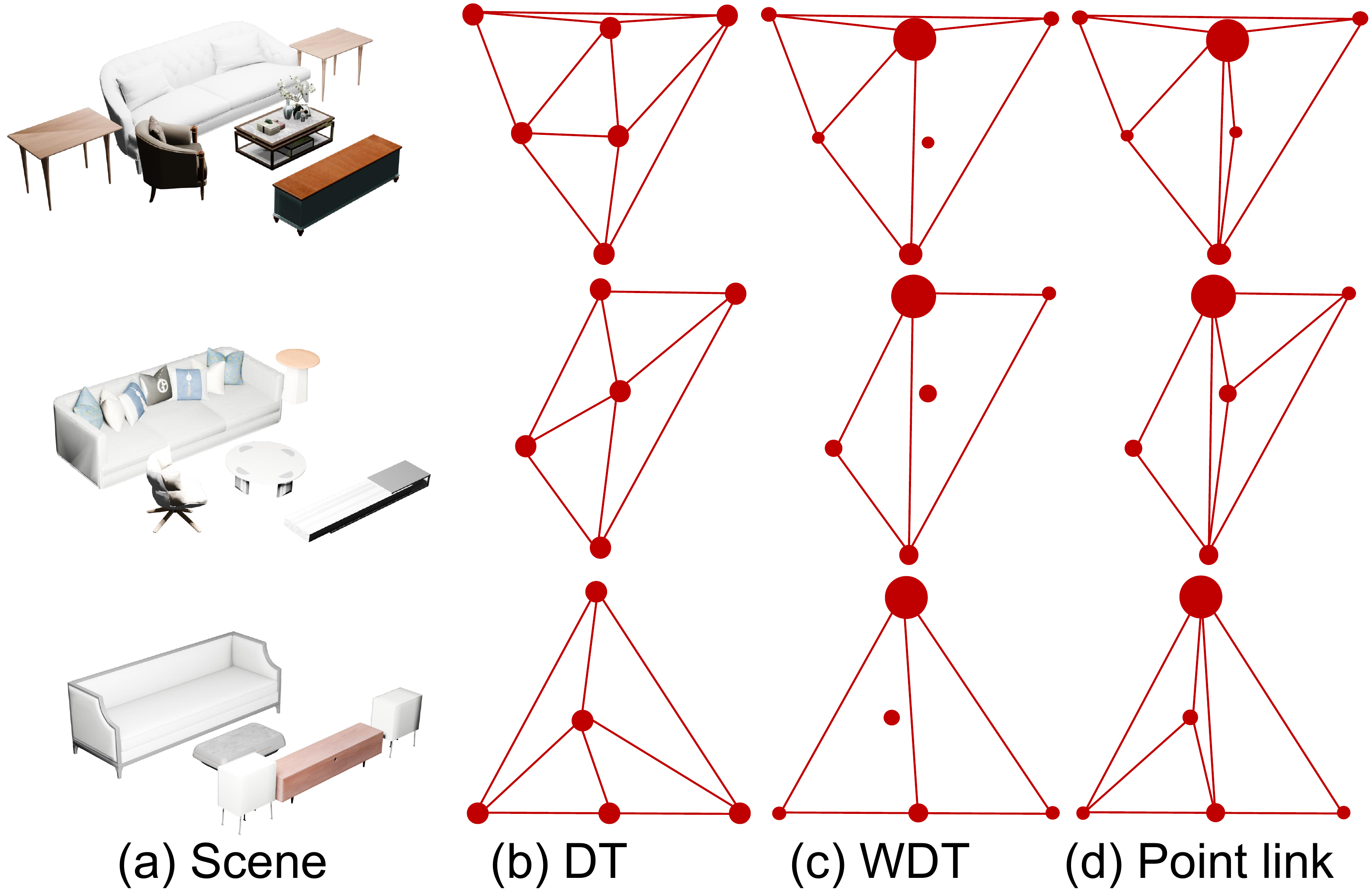}
    \caption{The comparison of the construction of O2O triplets using DT (b), WDT (c), and our PointLink (d) for input scenes (a).}
    \label{fig:o2o}  
\end{figure}

\noindent\textbf{O2O Triplets.} 
Each functional region usually has a functionally dominant object, e.g., a bed in a bedroom, and most other objects are affected by the position of the dominant object, e.g., chairs surrounding a dining table.  
Meanwhile, the proximity compatibility principle~\cite{wickens1995proximity} suggests that spatially close objects are more likely to be related. 
To fulfil these two criteria, we employ Weighted Delaunay Triangulation (WDT)~\cite{goes2014weighted} to partition object relationships into O2O triples with each object as a WDT point.  
As an object with larger occupancies is usually more important in a functional region, we calculate the WDT weight for each object using the multiplication of its size. 
Compared to Delaunay Triangulation (DT), WDT places more emphasis on larger objects and generates denser connections for large objects, as illustrated in Fig.~\ref{fig:o2o}. 
However, when two furniture objects exhibit significant weight disparity, WDT may fail to connect some points despite geometric proximity. Approximately $5\%$ of living rooms encountered such unconnected points.
To resolve this, we iterate through each triangle generated by the weighted Delaunay triangulation and link unconnected points and triangle points formed by WDT by performing DT inside each triangle with unconnected points, ensuring connectivity for all objects. 
If the region contains only two objects, we introduce an additional virtual node to construct an equilateral triplet. 
A comparative illustration of DT, WDT, and our approach is presented in Fig.~\ref{fig:o2o}.
Therefore, O2O relationships are converted to a set of O2O triplets among objects. 


\begin{figure}[htbp]
    \centering
    \includegraphics[width=0.46\textwidth]{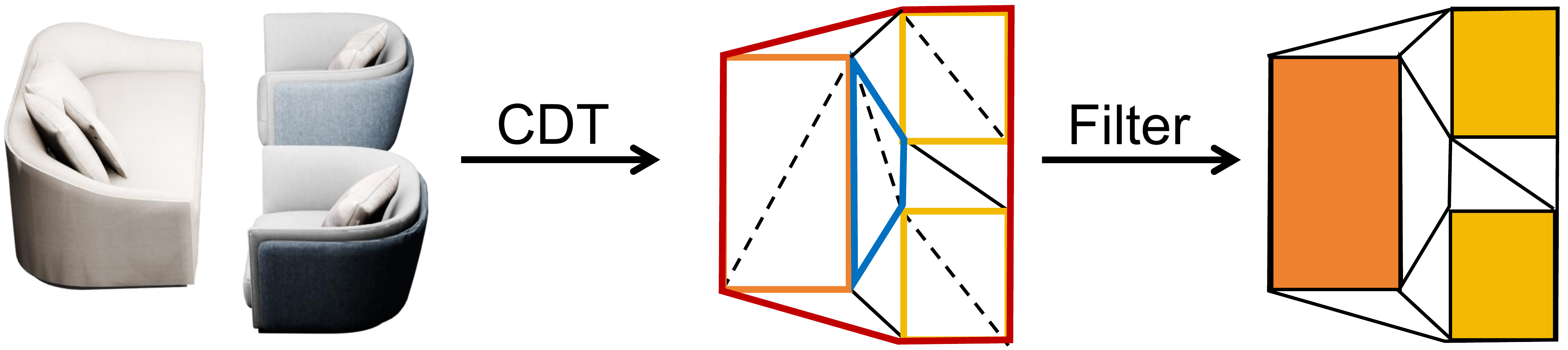}
    \caption{An example of the C2C triplet construction.}
    \label{fig:c2c}  
\end{figure}

\noindent\textbf{C2C Triplets.} 
As O2O triplets represent each object as a point and ignore fine-grained spatial relationships between the bounding boxes of two objects. We further introduce C2C triplets to account for this type of relationship. 
Fig.~\ref{fig:c2c} illustrates an example of constructing a C2C triplet. 
The minimum convex polygon formed by the vertices of three non-overlapping furniture bounding boxes is defined as the outer ring (shown in red in Fig.~\ref{fig:c2c}), while the remaining vertices constitute the inner ring (shown in blue in Fig.~\ref{fig:c2c}).
We tessellate the space between the inner ring and the outer ring with Constrained Delaunay Triangulation (CDT)~\cite{chew1987constrained}. 
The tessellated triangles formed by the corners of any two objects mark the relative spatial positions between two objects. 
The input to CDT is a planar straight-line graph with all corners as input vertices and the edges of object-triplet bounding boxes, the inner ring, and the outer ring as input line segments. 
We filter the resulting tessellated triangles that are within any bounding boxes and inner loops as the C2C triplets. 
These C2C triplets characterize the relative spatial positions between object corners. 


The relational triplets, including those derived from DBSCAN-based functional region grouping and DT methods, are precomputed for each training data and remain unchanged during training, regardless of the positions of their constituent objects perturbed by the noise injection. 
Therefore, they do not introduce additional computational overhead for the model training and have no impact on the denoising speed when performing as relational supervision as described in Sec.~\ref{subsec:loss}.

\subsection{Relationship Losses}\label{subsec:loss}


To encourage the layout diffusion model to capture hierarchical spatial relationships, we introduce relationship losses on the denoised layout $\bar{\mathbf{x}}_0^t$ during the training phase, namely $\mathcal{L}_\text{O2R}, \mathcal{L}_\text{O2O}, \mathcal{L}_\text{C2C}$, by measuring discrepancies between ground-truth O2R, O2O and C2C triplets and their denoised counterparts.
An intuitive approach is to supervise the geometric attributes of the triangles formed by relational triplets, which helps preserve their internal structure.
However, supervising multiple geometric attributes simultaneously is nontrivial, as their relative significance often shifts across different layouts and makes it difficult to strike a balance. 
Besides, we observe that a triplet is only functionally plausible when placed in an appropriate region of the layout.
For example, arbitrarily rotating an O2O triplet composed of a sofa and two corner-side tables in a living region, though not offending its internal geometry, can undermine the global layout not only due to this triplet being misaligned with the designated orientation, but also because it interferes with spatial relationships involved by other triplets. 

To this end, we expect a spatial-aware relationship loss that can effectively penalize spatial misalignment, while also offering a holistic perspective for capturing diverse geometric variations among triplets.
We intuitively employ a negative IoU loss. In order to avoid the instability of calculating triangles' intersection and for implementation convenience, we adopt differentiable rasterization of triangles and estimate IoU at the pixel level.
The loss $\mathcal{L}_\text{O2R}$ can be calculated as Eq.~\ref{Eq: L_O2R}:
\begin{equation}
\mathcal{L}_{\text{O2R}} = 1 -\frac{1}{|\overline{M}_\text{O2R}|}\cdot\sum_{\substack{\overline{m}_i \in \overline{M}_\text{O2R}, \\ m_i \in M_\text{O2R}}}\mathbf{IoU}(\mathcal{R}(\overline{m}_i),\mathcal{R}(m_i)),
\label{Eq: L_O2R}
\end{equation}
where $\mathcal{R}(\cdot)$ denotes differentiable rendering taking a triangle as input, $|\overline{M}_\text{O2R}|$ denotes the number of O2R triplets within the denoised layout $\bar{\mathbf{x}}_0^t$,
and $\mathbf{IoU}(\cdot, \cdot)$ computes the IoU between triangles formed by prediction and ground-truth triplets.

The O2O and C2C losses $\mathcal{L}_{\text{O2O}}, \mathcal{L}_{\text{C2C}}$ are computed using a similar formulation but with a slight modification: we align the centroids of the denoised and ground-truth triplets before computing the IoU.
This alignment ensures that the losses focus on structural differences in the triangle's shape and orientation, rather than being dominated by absolute spatial displacement.
In doing so, the loss better captures geometric discrepancies while remaining invariant to global translation.

The overall optimization objective is a summation $\mathcal{L} = \mathcal{L}_\text{diff} + \lambda_\text{O2R}\cdot\mathcal{L}_{\text{O2R}} + \lambda_\text{O2O}\cdot\mathcal{L}_{\text{O2O}} + \lambda_\text{C2C}\cdot\mathcal{L}_{\text{C2C}}$ with balancing weights $\lambda_\text{O2R}$, $\lambda_\text{O2O}$ and $\lambda_\text{C2C}$.


\section{experiments and results} \label{sec:exp}
In this section, we evaluate the method on three tasks, including unconditional indoor layout generation, floorplan-conditioned indoor layout generation, and indoor layout rearrangement. 
After that, we ablate key components and hyperparameters.   

\subsection{Experimental Settings}
\noindent\textbf{Datasets.} We use the 3D-FRONT dataset~\cite{fu20213dfront} for experiments. Following previous work~\cite{tang2024diffuscene,yang2024physcene}, we select a subset of rooms containing 4,041 bedrooms, 900 dining rooms, and 813 living rooms. 
For each room category, 80\% of the rooms are used for training, while the remaining 20\% are reserved for testing.
\begin{table}[htbp] 
\centering
\caption{Statistics of relative Euclidean distances between furniture pairs.}
\label{tab:furniture_statistics}
\renewcommand{\arraystretch}{1.0}
\setlength{\tabcolsep}{0.2pt} 
\resizebox{0.45\textwidth}{!}{ 
\begin{tabular}{llc} 
\hline
\textbf{Room}        & \textbf{Furniture Pair}            & \textbf{Average Distance}  \\ \hline
\textbf{Dining room} & cabinet - dining\_chair            & 0.524  \\
                     & dining\_chair - dining\_table      & 0.176  \\
                     & corner\_side\_table - dining\_table & 0.745  \\
                     & coffee\_table - dining\_table      & 0.689  \\
                     & coffee\_table - corner\_side\_table & 0.397  \\
                     & console\_table - dining\_chair     & 0.475  \\
                     & dining\_chair - dining\_chair      & 0.244  \\
                     & dining\_chair - multi\_seat\_sofa  & 0.734  \\
                     & armchair - dining\_chair           & 0.832  \\
                     & coffee\_table - dining\_chair      & 0.720  \\
                     & dining\_chair - tv\_stand          & 0.779  \\
                     & coffee\_table - tv\_stand          & 0.371  \\
                     & dining\_chair - loveseat\_sofa     & 0.795  \\
                     & corner\_side\_table - dining\_chair & 0.774  \\
                     & bookshelf - dining\_chair          & 0.541  \\
                     & dining\_chair - wine\_cabinet      & 0.459  \\
                     & dining\_chair - stool              & 0.526  \\ \hline
\textbf{Bedroom}     & double\_bed - wardrobe             & 0.730  \\
                     & nightstand - wardrobe              & 0.882  \\ 
                     & double\_bed - nightstand           & 0.571  \\ 
                     & nightstand - nightstand            & 0.907  \\ \hline
\textbf{Living room} & coffee\_table - multi\_seat\_sofa    & 0.256  \\
                     & corner\_side\_table - multi\_seat\_sofa & 0.315  \\
                     & armchair - coffee\_table           & 0.301  \\
                     & armchair - corner\_side\_table     & 0.418  \\
                     & coffee\_table - corner\_side\_table & 0.405  \\
                     & coffee\_table - dining\_table      & 0.689  \\
                     & dining\_chair - dining\_table      & 0.173  \\
                     & dining\_chair - dining\_chair      & 0.231  \\
                     & dining\_chair - loveseat\_sofa     & 0.784  \\
                     & coffee\_table - tv\_stand          & 0.366  \\
                     & corner\_side\_table - tv\_stand    & 0.638  \\
                     & coffee\_table - dining\_chair      & 0.719  \\
                     & dining\_table - tv\_stand          & 0.723  \\
                     & corner\_side\_table - dining\_table & 0.735  \\
                     & dining\_chair - tv\_stand          & 0.757  \\
                     & corner\_side\_table - dining\_chair & 0.767  \\
                     & bookshelf - dining\_chair          & 0.526  \\
                     & dining\_chair - multi\_seat\_sofa  & 0.742  \\
                     & armchair - dining\_chair           & 0.822  \\
                     & dining\_chair - wine\_cabinet      & 0.493  \\
                     & console\_table - dining\_chair     & 0.483  \\
                     & dining\_chair - stool              & 0.571  \\
                     & cabinet - dining\_chair            & 0.568  \\
                     & dining\_chair - lounge\_chair      & 0.797  \\ \hline
\end{tabular}
}
\end{table}

\begin{table*}[htbp]
\centering
\small
\caption{Quantitative comparisons on unconditioned layout generation.}
\label{tab:uncond}
\resizebox{0.95\textwidth}{!}{ 
\begin{tabular}{lcccccccc}  
\toprule  
    Room & Method & FID$\downarrow$ & KID{\tiny{}$\times0.001$}$\downarrow$ & SCA & CKL{\tiny{}$\times0.01$}$\downarrow$ & Col\textsubscript{obj}$\downarrow$ & Col\textsubscript{scene}$\downarrow$ & D\textsubscript{obj}$\downarrow$ \\  
\midrule  
    \multirow{3}{*}{Bedroom} & ATISS & 56.628 & 4.764 & 0.542 & 0.752 & 0.291 & 0.546 & 0.172 \\
    & LayoutEnhancer & 42.515& 4.241& 0.523& 0.526& 0.256& 0.488& 0.108\\ 
    & DiffuScene & 39.422& 4.352& 0.527& 0.396& 0.247& 0.462& 0.072\\  
    & Forest2Seq & 38.412& 3.925& 0.519& 0.343& 0.238& 0.457& 0.062\\ 
    \cline{2-9}
    & Ours &\textbf{27.247} &\textbf{1.281}  &\textbf{0.511}& \textbf{0.254}&\textbf{0.212}&\textbf{0.410}  &\textbf{0.030}  \\
\hline  
    \multirow{3}{*}{Living room} & ATISS & 74.522 & 5.671 & 0.592 & 0.511 & 0.384 & 0.878 & 0.672 \\  
    & LayoutEnhancer & 68.462& 4.862& 0.577& 0.518& 0.314& 0.748& 0.613\\ 
    & DiffuScene & 50.699& 3.792& 0.562& 0.234& 0.221& 0.612& 0.432\\  
    & Forest2Seq & 48.235& 1.894& 0.529& 0.231& 0.219& 0.589& 0.398\\ 
    \cline{2-9}
    & Ours  &\textbf{44.566}  &\textbf{0.711}  &\textbf{0.496}  &\textbf{0.125}  &\textbf{0.135}   &\textbf{0.449}   & \textbf{0.206} \\
\hline  
    \multirow{3}{*}{Dining room} & ATISS & 72.433 & 5.524 & 0.586 & 0.653 & 0.422 & 0.806 & 0.581 \\  
    & LayoutEnhancer & 69.125& 4.924& 0.545& 0.529& 0.352& 0.698& 0.492\\ 
    & DiffuScene & 49.658& 4.256& 0.538& 0.231& 0.224& 0.628& 0.423\\   
    & Forest2Seq & 49.127& 2.567& 0.531& 0.229& 0.202& 0.572& 0.386\\ 
    \cline{2-9}
    & Ours &\textbf{46.453}&\textbf{2.434}&\textbf{0.509}  &\textbf{0.161}  &\textbf{0.160} &\textbf{0.530} &\textbf{0.185} \\
\bottomrule 
\end{tabular}}
\end{table*}

\begin{figure*}[htbp]
   \centering
   \includegraphics[width=0.9\textwidth]{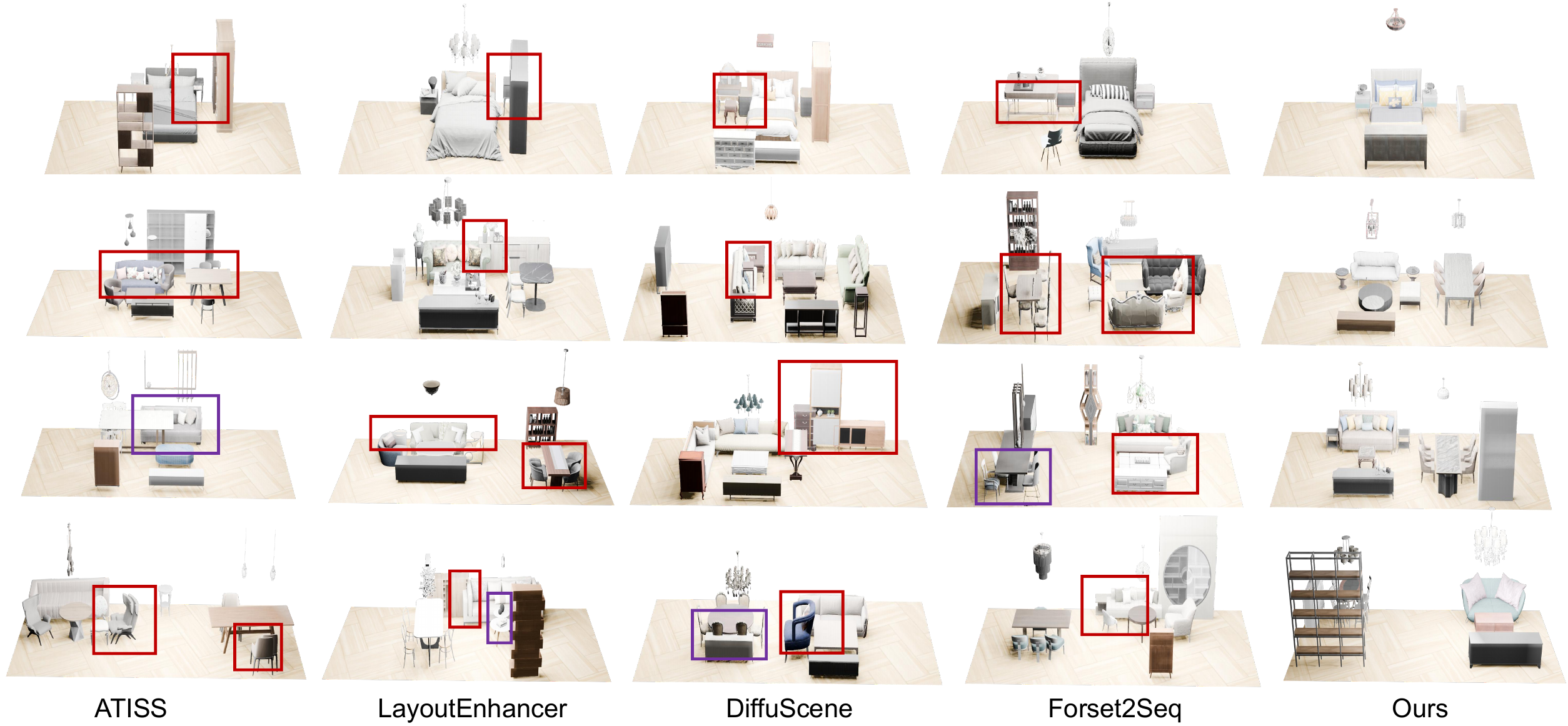}
   \caption{Qualitative results on unconditional layout synthesis. The \textcolor{red}{red} boxes indicate overlaps, while the \textcolor{violet}{purple} boxes highlight unreasonable furniture distribution.}
   \label{fig:unconditional_scene_comparison}
\end{figure*}

\noindent\textbf{Metrics.} To evaluate the visual quality of generated scenes, we follow Diffuscene~\cite{tang2024diffuscene} and employ perceptual similarity Fréchet Inception Distance (FID)~\cite{heusel2017gans}, Kernel Inception Distance (KID × 0.001)~\cite{BinkowskiSAG18}, Scene Classification Accuracy (SCA), and categorical KL Divergence (CKL × 0.01).
FID, KID, and SCA are highly sensitive to shape textures and rendering styles. To ensure fairness, we render the results of all competing methods using the same configuration.
Following PhyScene~\cite{yang2024physcene}, we adopt {Col\textsubscript{obj}}, {Col\textsubscript{scene}}, and {R\textsubscript{out}} to assess the violation of important spatial relationships.
{Col\textsubscript{obj}} represents the percentage of objects that collide with others, and {Col\textsubscript{scene}} denotes the percentage of generated layouts where the object collision occurs. 
{R\textsubscript{out}} evaluates the proportion of objects that extend beyond the floorplan in the floor-conditioned layout synthesis. 
For {Col\textsubscript{obj}}, {Col\textsubscript{scene}}, {R\textsubscript{out}}, lower values are better.

We additionally introduce a metric D\textsubscript{obj} to measure the changes in spatial distance distribution between frequently co-occurring object pairs in synthesised layouts and ground truth.
A furniture pair is frequent if it appears more than a specified portion ($50\%$ in our experiments). 
For all furniture pairs with specific semantics in the training set, we compute the Euclidean distance between their normalized centroids, and adopt Gaussian kernel density estimation for their distance distribution, with the bandwidth for the kernel selected via Scott's rule~\cite{Scott1992MultivariateDE}.
At last, D\textsubscript{obj} takes the averaged absolute differences of distance expectations between predicted furniture pairs and ground truth ones as the metric. 
The distance expectation for all frequent furniture pairs in the ground truth is presented in Tab.~\ref{tab:furniture_statistics}.

\noindent\textbf{Baselines.} For indoor layout synthesis, we compare with four state-of-the-art methods: 1) ATISS~\cite{paschalidou2021atiss}, an autoregressive network for indoor layout generation; 2) LayoutEnhancer~\cite{leimer2022layoutenhancer}, an autoregressive model incorporating ergonomic and indoor functional design knowledge as a training regularization term; 3) DiffuScene~\cite{tang2024diffuscene}, a DDPM-based indoor layout synthesis model that introduces IoU regularization terms to penalize object collisions; 4) PhyScene~\cite{yang2024physcene}, which models pairwise object relationships as a binary classifier and provides classifier guidance for a DDPM-based indoor layout synthesis model;
and 5) Forest2Seq~\cite{sun2024forest2seq}, a method that organizes unordered scene objects into a structured and ordered hierarchical scene tree and forest.
Since PhyScene only releases the inference code and the pre-trained model for a floor-conditioned living room. We compare with PhyScene only under this configuration.
In addition, we also compare the method with DiffuScene and LEGO~\cite{wei2023lego} for scene rearrangement tasks. 

\begin{table*}[htbp]
    \centering
    \small
\caption{Quantitative comparisons of floor-conditioned layout synthesis across different methods.}
    \label{tab:Floor-conditioned}
\resizebox{0.95\textwidth}{!}{ 
\begin{tabular}{lccccccccccc}  
\toprule  
    Room & Method & FID$\downarrow$ & KID{\tiny{}$\times0.001$}$\downarrow$ & SCA & CKL{\tiny{}$\times0.01$}$\downarrow$ & Col\textsubscript{obj}$\downarrow$ & Col\textsubscript{scene}$\downarrow$ & D\textsubscript{obj}$\downarrow$ & R\textsubscript{out}$\downarrow$ \\  
\midrule  
    \multirow{5}{*}{Bedroom} & ATISS & 50.871 & 2.264 & 0.533 & 0.323 & 0.278 & 0.593 & 0.102 & 0.302 \\  
    & LayoutEnhancer & 48.984 & 2.143 & 0.525 & 0.299 & 0.247 & 0.465 & 0.082 & 0.276 \\  
    & DiffuScene & 37.060 & 2.163 & 0.528 & 0.312 & 0.223 & 0.434 & 0.056 & 0.268 \\  
    & Forest2Seq & 36.102 & 2.125 & 0.521 & 0.268 & 0.215 & 0.408 & 0.049 & 0.242 \\  
    \cline{2-10}
    & Ours &\textbf{28.910} &\textbf{1.901} &\textbf{0.516} &\textbf{0.221}  &\textbf{0.181}  &\textbf{0.329}&\textbf{0.036}   & \textbf{0.205} \\  
\hline  
    \multirow{5}{*}{Living room} & ATISS & 62.920 & 4.768 & 0.541 & 0.212 & 0.316 & 0.876 & 0.771 & 0.152 \\  
    & LayoutEnhancer & 61.520 & 4.252 & 0.535 & 0.208 & 0.285 & 0.874 & 0.652 & 0.149 \\  
    & DiffuScene & 49.421 & 3.546 & 0.544 & 0.203 & 0.254 & 0.706 & 0.664 & 0.248 \\  
    & PhyScene & 50.361 & 3.632 & 0.539 & 0.171 & 0.202 & 0.630 & 0.629 & 0.220\\  
    & Forest2Seq & 48.998 & 2.989 & 0.538 & 0.189 & 0.217 & 0.654 & 0.649 & \textbf{0.147} \\  
    \cline{2-10}
    & Ours &\textbf{46.335}&\textbf{1.854}  & \textbf{0.514} &\textbf{0.160}   &\textbf{0.172}  &\textbf{0.499}&\textbf{0.482}&0.155  \\  
\hline  
    \multirow{5}{*}{Dining room} & ATISS & 65.136 & 4.762 & 0.520 & 0.316 & 0.415 & 0.860 & 0.571 & 0.146 \\  
    & LayoutEnhancer & 60.106 & 3.882 & 0.518 & 0.298 & 0.323 & 0.824 & 0.480 & 0.143 \\  
    & DiffuScene & 48.908 & 2.889 & 0.517 & 0.204 & 0.185 & 0.586 & 0.455 & 0.202 \\  
    & Forest2Seq & 48.312 & 2.792 & 0.515 & 0.191 & 0.181 & 0.579 & 0.428 & 0.139 \\  
    \cline{2-10}
    & Ours &\textbf{46.799}  &\textbf{2.490}   &\textbf{0.510}   &\textbf{0.151}  &\textbf{0.148}  &\textbf{0.479}  &\textbf{0.363}  &\textbf{0.132}  \\  
\bottomrule  
\end{tabular}}
\end{table*}

\begin{figure*}[htbp]
    \centering
    \includegraphics[width=\linewidth]{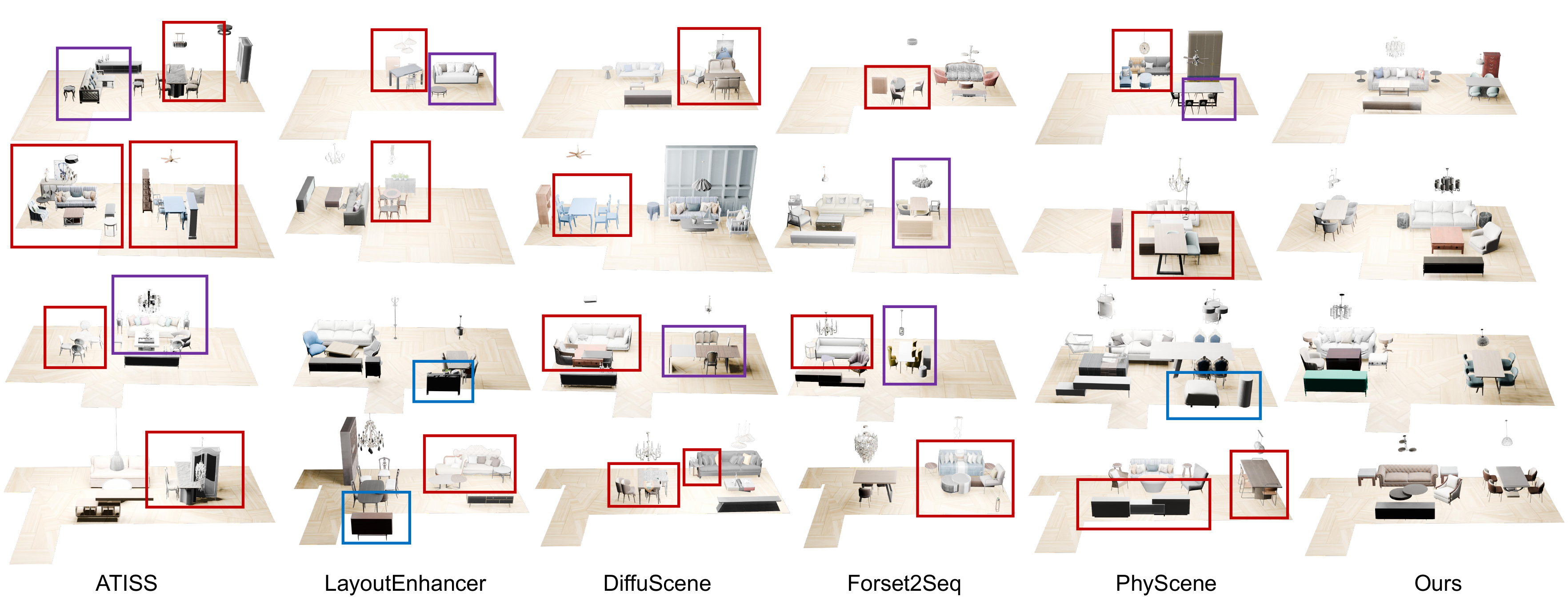}
        \caption{Visualization comparisons of floor-plan conditioned layout synthesis. The \textcolor{red}{red} boxes indicate overlap, the \textcolor{blue}{blue} boxes indicate out-of-bounds, and the \textcolor{violet}{purple} boxes highlight areas with unreasonable furniture distribution.}
    \label{fig:conditioned}  
\end{figure*}

\noindent\textbf{Implementation} We train our model using a single NVIDIA A100 GPU, with a batch size of 128, for 100, 000 epochs. The initial learning rate was set at $2\times10^{-4}$ and gradually decayed by a factor of 0.5 for every 15,000 epochs. For the diffusion process, we follow the default settings of DDPM~\cite{ho2020denoising}, where the noise intensity $\beta_t$ increases linearly from $0.0001$ to $0.02$ for $1,000$ time steps. The complete training for the bedroom, living room, and dining room takes three days. For DBSCAN clustering, epsilon and the min number of samples are set to 1.8 and 2, respectively, with a GIoU weight of $\lambda_{\text{giou}}=0.02$. During training, we set the loss weight $\lambda_{\text{O2R}}=0.025, \lambda_{\text{O2O}}=0.05, \lambda_{\text{C2C}}=0.04$.
The output images are rendered with dimensions of 128 pixels by 128 pixels.
To evaluate visual metrics FID, KID, and SCA, we construct 3D scenes from the generated indoor layout by retrieving the closest matched CAD furniture from 3D-FUTURE \cite{fu20213dfuture} based on the 3D object feature $f_i$. 

\subsection{Unconditioned Indoor Layout Generation} 
In this scenario, the method directly generates layouts without any input conditions.  
The quantitative results are presented in Tab.~\ref{tab:uncond}. 
\method{} demonstrates a clear advantage across visual quality metrics (FID, KID, SCA, and CKL), showing consistently superior generation quality compared to ATISS, LayoutEnhancer, and DiffuScene. These metrics underscore the effectiveness of incorporating spatial relationships for perceptual and semantic accuracy, which is crucial for generating realistic layouts.
In terms of collision metrics, \method{} significantly reduces object overlaps and boundary violations. 
For the living room, HierRelTriple achieves reductions of 38.4\% in Col\textsubscript{obj} and 23.8\% in Col\textsubscript{scene}, relative to Forest2Seq.
This improvement indicates effective collision management, suggesting that \method{} is better at collision.

Additionally, \method{} achieves superior performance in D\textsubscript{obj}, highlighting its ability to maintain consistent distance relationships. In the bedroom layout, for example, \method{} achieves a D\textsubscript{obj} of 0.030, compared to Forest2Seq (0.062) and DiffuScene (0.072). This reflects \method{}'s precision in preserving relative spacing, contributing to both visual authenticity and physical rationality within generated scenes.

Fig.~\ref{fig:unconditional_scene_comparison} presents a qualitative comparison of different layout generation methods. Our generated layouts have fewer object collisions, and the object placement is more reasonable, for example, the proper spacing between a bed and nightstands, a sofa, and its surrounding objects. 

\begin{figure*}[htbp]
    \centering
    \includegraphics[width=0.98\textwidth]{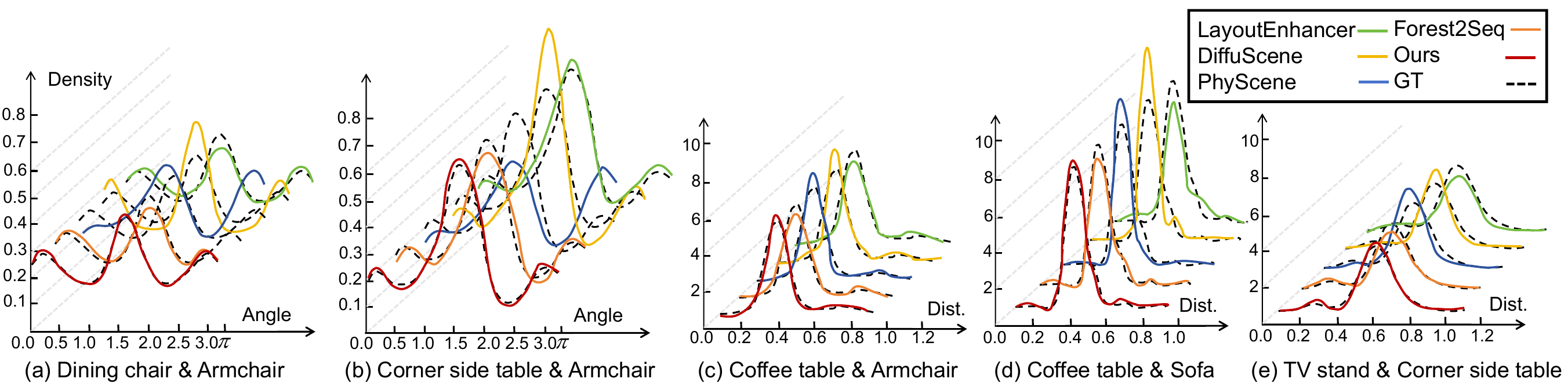}
    \caption{Distribution of the relative distances and angles between object pairs modeled by five approaches—LayoutEnhancer, DiffuScene, PhyScene, Forest2Seq, and Ours—compared to the ground truth, where (a)-(b) visualizes the angle distribution and (c)-(e) visualizes the distance distribution.}
    \label{fig:distance_comparison} 
\end{figure*}

\subsection{Floorplan-conditioned Layout Generation}
In this scenario, the method generates layouts conditioned on a floorplan input.  
Tab.~\ref{tab:Floor-conditioned} compares \method{} with alternative baselines.
Note that PhyScene only provided the pre-trained model and testing code for living room comparisons.
Across metrics that assess layout realism and spatial organization, \method{} demonstrates distinct advantages over DiffuScene, LayoutEnhancer, PhyScene, and Forest2Seq.  For the living room layout, for example, \method{} achieves a D\textsubscript{obj} of 0.482, much lower than DiffuScene(0.664) and PhyScene(0.629). This lower $D_{obj}$ value indicates \method{}’s superior control over object distances, effectively maintaining realistic spacing that contributes to both visual coherence and physical plausibility in the layout.

\method{}’s R\textsubscript{out} is marginally higher than Forest2Seq. A possible reason is that \method{} emphasizes the relationship between objects and regions rather than room boundaries, leading to higher R\textsubscript{out} but significant improvements in D\textsubscript{obj}, Col\textsubscript{obj}, and Col\textsubscript{scene} metrics.
Fig.~\ref{fig:conditioned} illustrates a visual comparison between various layout synthesis methods.
In Fig.~\ref{fig:distance_comparison}, we provide a kernel density estimate (KDE) plot of the distance distribution between co-occurring furniture pairs. 
We can see our method is much closer to the ground truth distribution at the highest density peak and can accurately track the trends of the overall density changes. 
Therefore, the generated layouts exhibit more similar distance distributions to the ground truth layouts than other methods. 
Our method produces fewer object obstructions and has less unoccupied space than Forest2Seq and PhyScene.

\subsection{Indoor Layout Re-arrangement}
\begin{figure}[htbp]
    \centering
    \includegraphics[width=\linewidth]{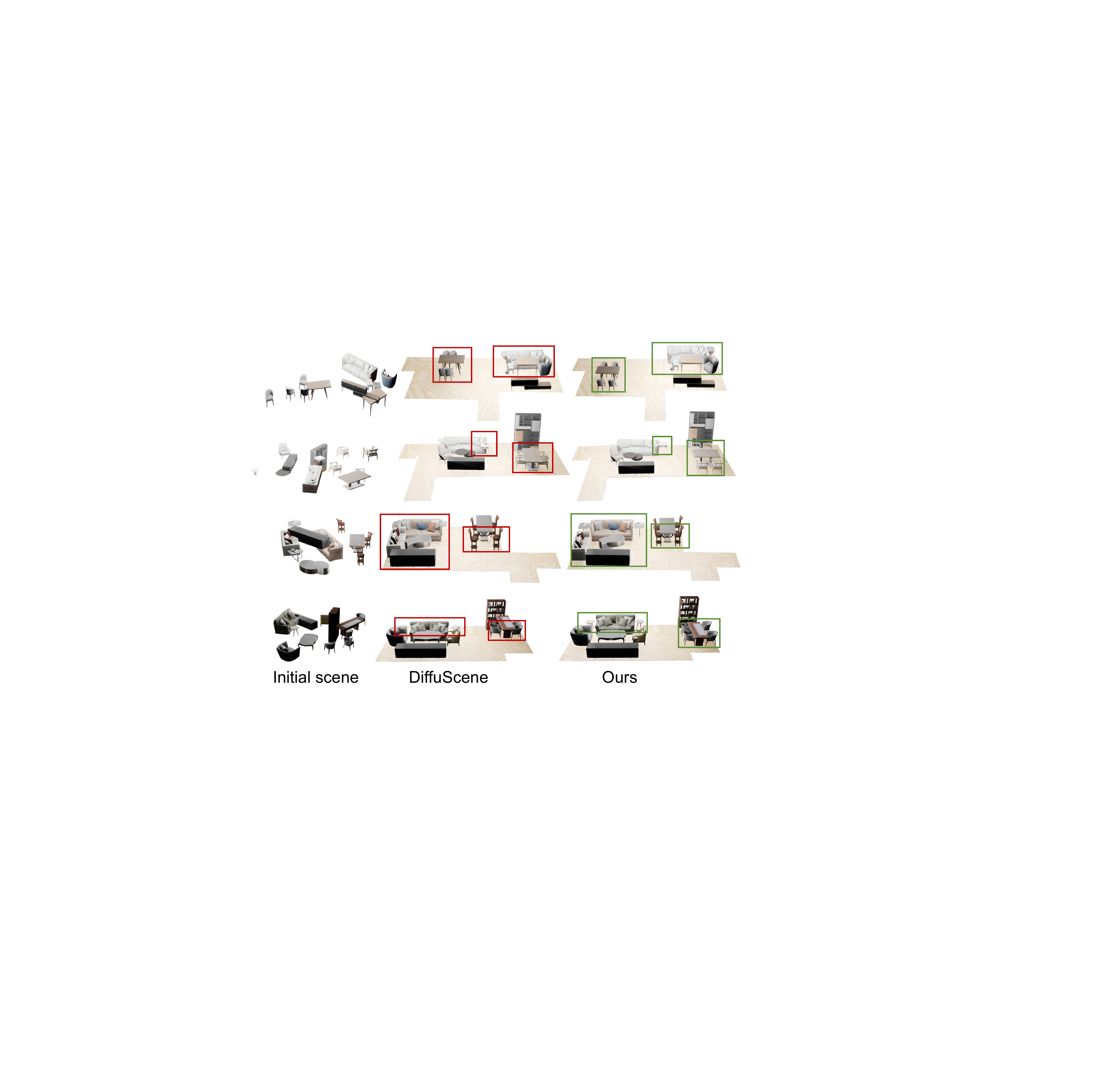}
        \caption{Visual comparisons on scene re-arrangements where the indoor objects are given. The \textcolor{red}{red} boxes mark incorrect furniture placements, while the \textcolor{green!50!black}{green} boxes indicate appropriate arrangements. }
    \label{fig:rearranged}  
\end{figure}

In this task, we take a set of semantic objects and generate an indoor layout for the living room setting. 
Tab.~\ref{tab:re-arrangement} presents the results. 
We can see the inclusion of \method{} can greatly improve the performance of both baselines and achieve better numbers on all metrics, especially for constraint metrics (see improvements in Col\textsubscript{obj}, Col\textsubscript{scene}, and D\textsubscript{obj}). 
Qualitative results in Fig.~\ref{fig:rearranged} show superior performance in avoiding object collisions. 
The integration of \method{} leads to more realistic and functional indoor layouts by enhancing constraint satisfaction.
\begin{table}[htbp]
\caption{Quantitative comparisons on the task of scene arrangement in the 3D-FRONT living rooms. We evaluate results with
LEGO and DiffuScene as baseline methods. }
\label{tab:re-arrangement}
\centering
\setlength{\tabcolsep}{2pt}
\renewcommand{\arraystretch}{1.5}  
\resizebox{0.45\textwidth}{!}{
\begin{tabular}{lcccccc}  
\toprule  
    Method & FID$\downarrow$ & KID{\tiny{}$\times0.001$}$\downarrow$ & SCA & Col\textsubscript{obj}$\downarrow$ & Col\textsubscript{scene}$\downarrow$ & D\textsubscript{obj}$\downarrow$ \\  
\midrule  
    LEGO &66.952 &7.592 &0.592 &0.557 &0.898 &0.725\\
    Ours(LEGO) &\textbf{54.912}  &\textbf{3.377}  &\textbf{0.567}  &\textbf{0.386}  &\textbf{0.796}  &\textbf{0.312}  \\
    \hline
    Diffuscene &61.017 &5.842 &0.586 &0.497 &0.920 &0.495\\ 
    Ours(Diffuscene) &\textbf{52.689} &\textbf{2.796} &\textbf{0.559} &\textbf{0.298} &\textbf{0.687} &\textbf{0.278} \\
\bottomrule  
\end{tabular}}
\end{table}

\begin{table*}[htbp]
    \centering
    \small
    \caption{Analysis of alternative spatial relationship modeling strategies.}
    \label{tab:relationship}
    \resizebox{0.95\textwidth}{!}{ 
    \renewcommand{\arraystretch}{1.5} 
    \begin{tabular}{cc|cccccccccc}
    \toprule
    spt. relationship & reg. term   & FID$\downarrow$ & KID{\tiny{}$\times0.001$}$\downarrow$ &  SCA & CKL{\tiny{}$\times0.01$}$\downarrow$ & Col\textsubscript{obj}$\downarrow$ &  Col\textsubscript{scene}$\downarrow$  & D\textsubscript{obj}$\downarrow$ \\ \midrule
    pairwise &  --    &46.184 &1.026 &0.516 &0.226 &0.191&0.569 &0.400 \\ \midrule
    triplet 
    & MSE-based loss &45.860 &0.950 &0.511 &0.189 &0.142 &0.510 &0.217 \\ 
    & IoU-based loss &\textbf{44.566}  &\textbf{0.711}  &\textbf{0.496}  &\textbf{0.125}  &\textbf{0.135}   &\textbf{0.449}   & \textbf{0.206}  \\ 
    \bottomrule
    \end{tabular}
}
\end{table*}

\subsection{Ablation Studies} \label{subsec:ablation}
In this section, we first evaluate the impact of different triangulation methods on the O2O component. Next, we assess the effectiveness of the three proposed loss functions. Finally, we compare our method with four alternative rendering loss designs. All experiments are conducted for unconditional living room generation.

\noindent\textbf{Alternatives for O2O Triplet Construction.}  We compared the impact of different triangulation methods: selecting two nearest neighbors for each object as O2O triplets (2NN), replacing WDT with DT, WDT, and our WDT with point links on the generated results. The results presented in Tab.~\ref{tab:o2o_ablation} demonstrate that WDT with point links yields the best performance, indicating O2O triplets constructed with an emphasis on object sizes can effectively identify important relationships.

\begin{table}[htbp]
\centering
\caption{Ablation study on triangulation methods for O2O.}
\label{tab:o2o_ablation}
\setlength{\tabcolsep}{1pt} 
\renewcommand{\arraystretch}{1.5}  
\begin{tabular}{lccccccc}
\toprule
Method & 
\multicolumn{1}{c}{FID$\downarrow$} & 
\multicolumn{1}{c}{KID{\tiny{}$\times0.001$}$\downarrow$} & 
\multicolumn{1}{c}{SCA} & 
\multicolumn{1}{c}{CKL{\tiny{}$\times0.01$}$\downarrow$} & 
\multicolumn{1}{c}{Col\textsubscript{obj}$\downarrow$} & 
\multicolumn{1}{c}{Col\textsubscript{scene}$\downarrow$} & 
\multicolumn{1}{c}{D\textsubscript{obj}$\downarrow$} \\
\midrule
2NN &46.962 &1.124 &0.521  &0.212  &0.154  &0.532 &0.352 \\
DT &45.860 &0.950  &0.511  &0.189  &0.142  &0.510  &0.217  \\
WDT &44.912  &0.836  &0.507  &0.138  &0.139  &0.460  &0.211  \\
Point links & \textbf{44.566} & \textbf{0.711} & \textbf{0.496} & \textbf{0.125} & \textbf{0.135} & \textbf{0.449} & \textbf{0.206} \\
\bottomrule
\end{tabular}
\end{table}

\noindent\textbf{Impact of O2R, O2O, and C2C Losses.} In Tab.~\ref{tab:ablation}, we explore how O2R and O2O losses affect the generation quality.
We evaluate the results of the unconditional generation in living rooms.
The O2R loss focuses on fitting objects within designated regions of the indoor layout, ensuring that each object is placed appropriately within the layout. 
Comparing row 1 and row 2, the generated indoor layout also achieves significant improvements in $\text{Col}_\text{obj} \ (0.254 \rightarrow 0.168)$, $\text{Col}_\text{scene} \ (0.762 \rightarrow 0.555)$, and $\text{D}_\text{obj} \ (0.452 \rightarrow 0.292)$.
Comparing row 1 and row 3 in Tab.~\ref{tab:ablation}, the O2O loss plays a crucial role in managing important spatial relationships ($\text{Col}_\text{obj} \ 0.254 \rightarrow 0.140$, $\text{Col}_\text{scene} \ 0.762 \rightarrow 0.460$), and shows better distribution of object spacing ($\text{D}_\text{obj} \ 0.452 \rightarrow 0.218$), indicating the avoidance of overcrowding or excessive spacing between objects. 
The C2C loss enhances the learning of spatial relationships. By comparing row 1 and row 4 in Tab.~\ref{tab:ablation}, the object spacing ($\text{D}_\text{obj} \ 0.452 \rightarrow 0.215$) is better regulated.
Combining all three hierarchical relationship losses achieves the best values on all metrics. 

\begin{table}[htbp]
\centering
    \caption{Ablation study on the relationship losses.}
\label{tab:ablation}
\setlength{\tabcolsep}{0.5pt} 
\renewcommand{\arraystretch}{1.5}  
\resizebox{\linewidth}{!}{
\begin{tabular}{ccccccccccc}
\toprule
  O2R & O2O & C2C & FID$\downarrow$ & KID{\tiny{}$\times0.001$}$\downarrow$ &  SCA & CKL{\tiny{}$\times0.01$}$\downarrow$ & Col\textsubscript{obj}$\downarrow$ &  Col\textsubscript{scene}$\downarrow$  & D\textsubscript{obj}$\downarrow$ \\

\midrule
 & &  & 51.462 & 3.925 & 0.579 & 0.273 & 0.254 &  0.762 & 0.452 \\
\checkmark& & &46.469 &1.119 &0.519 &0.238 &0.168 &0.555 &0.292 \\
 &\checkmark & &45.564 &0.868 &0.495 &0.151 &0.140 &0.460 &0.218 \\
 & &\checkmark &45.722 &0.862 &0.507 &0.208 &0.156& 0.523&0.215 \\
\hline
\checkmark & \checkmark & \checkmark &\textbf{44.566}  &\textbf{0.711}  &\textbf{0.496}  &\textbf{0.125}  &\textbf{0.135}   &\textbf{0.449}   & \textbf{0.206} \\
\bottomrule
\end{tabular}
}
\end{table}

\begin{figure}[htbp]
    \centering
    \includegraphics[width=0.9\linewidth]{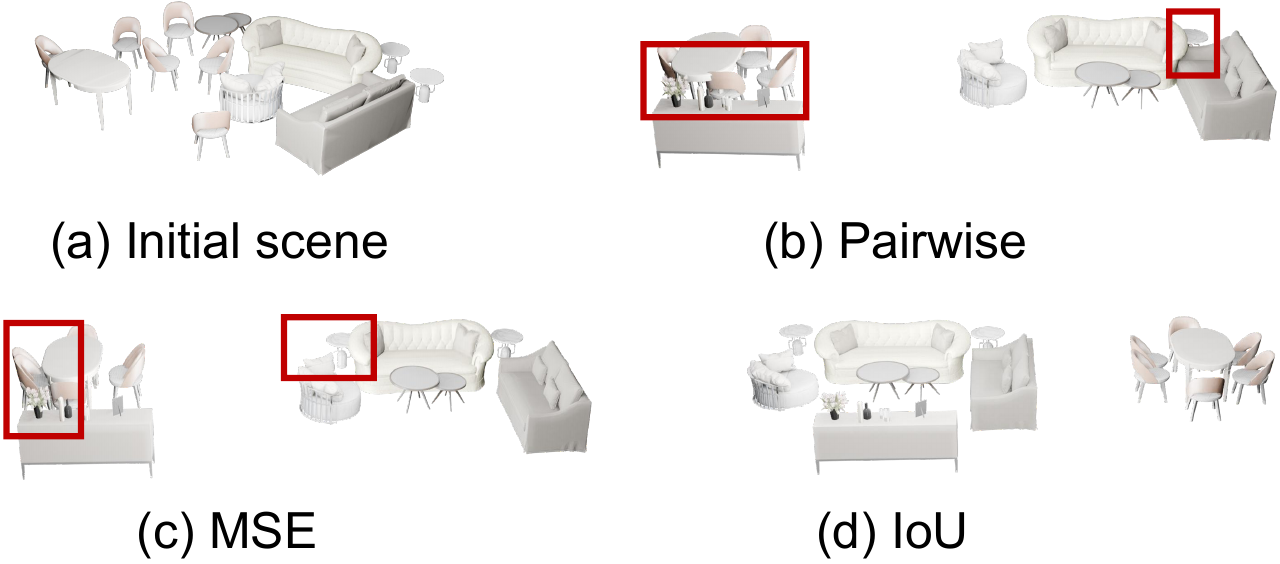}
        \caption{Comparative visualization of same-scene rearrangement with different relationship-modeling regularizations. The \textcolor{red}{red} boxes highlight incorrect furniture placements.}
    \label{fig:spatial_relationship_modeling}  
\end{figure}

\noindent\textbf{Alternatives to the IoU-based Relationship Losses.}
We validate the IoU-based loss against two other alternatives: 1) pairwise distance-based loss, which minimizes the pairwise distance variation between two objects in the triplet when comparing predictions with the ground truth (GT); 2) rendering the triplets and calculating the pixel-level mean squared error (MSE). The visualization results of scene rearrangement using different losses are shown in Fig.~\ref{fig:spatial_relationship_modeling}. The results in Tab.~\ref{tab:relationship} demonstrate that the best performance is achieved by calculating the IoU-based loss.

\subsection{Analysis and Discussions}\label{subsec:analysis}
\noindent\textbf{Rotation and Translation Invariance.}
We first validate whether rotation invariance can improve the results when comparing the predicted object triplet and the corresponding GT object triplet. 
For the rotation invariance setting, we first align the edge with the same semantic labels and the longest length from the two comparing triplets, and then calculate the IoU loss. 
As shown in Tab.~\ref{tab:invariance}, introducing rotation invariance yields no performance improvement.

Regarding translation invariance, we compared two triplet configurations: one centered at the origin (introducing translation invariance) and one without centering (removing translation invariance). Results in Tab.~\ref{tab:invariance} demonstrate that incorporating translation invariance improves performance. 

\begin{figure*}[htbp]
    \centering
    \includegraphics[width=0.9\textwidth]{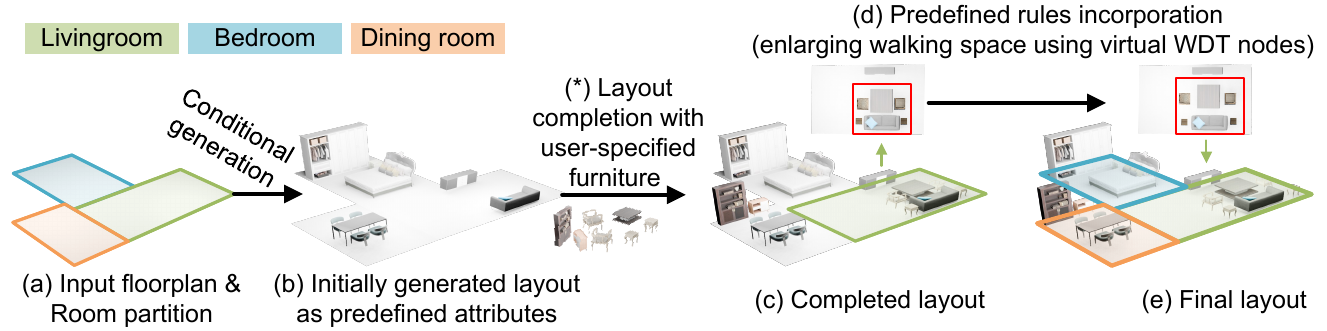}
    \caption{A workflow example for layout generation and editing with predefined attributes and rules.}
    \label{fig:application} 
\end{figure*}
\begin{figure}[htbp]
    \centering
    \includegraphics[width=\linewidth]{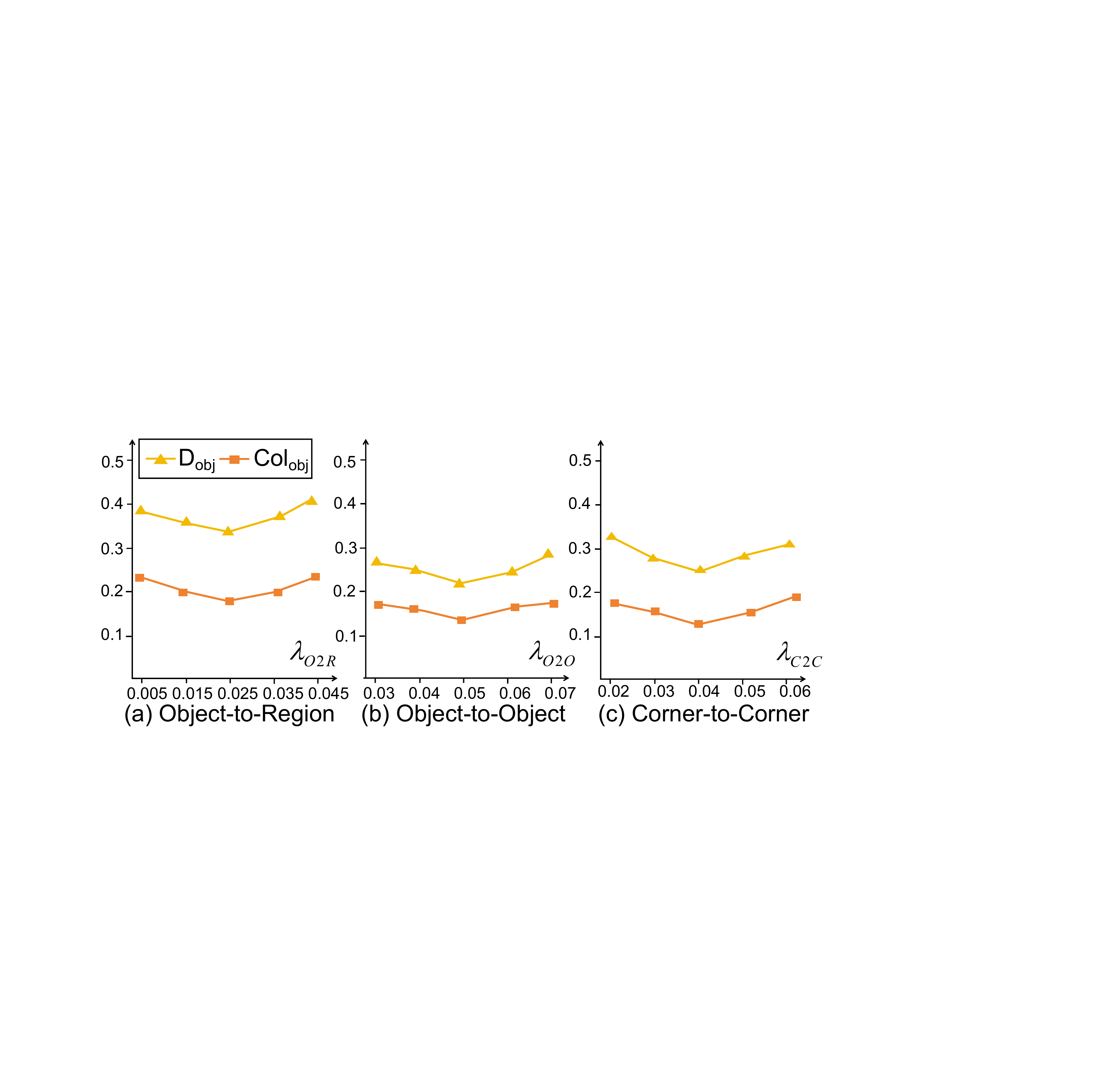}
    \caption{Evaluations on the varying weights for loss balancing.}
    \label{fig:HyperParameters}
\end{figure}
\begin{table}[htbp]
    \centering
    \small
    \caption{The impact of translation and rotation invariance on the performance.}
    \label{tab:invariance}
    \setlength{\tabcolsep}{0.3pt} 
    \renewcommand{\arraystretch}{1.5}  
    \resizebox{0.5\textwidth}{!}{ 
    \begin{tabular}{c|cccccccccc}
    \toprule
    Invariance & FID$\downarrow$ & KID{\tiny{}$\times0.001$}$\downarrow$ & SCA & CKL{\tiny{}$\times0.01$}$\downarrow$ & Col\textsubscript{obj}$\downarrow$ & Col\textsubscript{scene}$\downarrow$  & D\textsubscript{obj}$\downarrow$ \\ \midrule
    None & 46.291 &1.466 &0.511 &0.213 &0.159&0.531&0.219  \\
    Rotation &47.136 &1.423  &0.512 &  0.213&0.162 &0.535 &0.228\\
    Translation &\textbf{44.566}  &\textbf{0.711}  &\textbf{0.496}  &\textbf{0.125}  &\textbf{0.135}   &\textbf{0.449}   & \textbf{0.206} \\
    \bottomrule
    \end{tabular}
    }
\end{table}

\noindent\textbf{Weights for Loss Functions.}
We validate the balancing weights for O2O, O2R, and C2C relationships with respect to the D\textsubscript{obj} and Col\textsubscript{obj} metrics. 
For the O2R loss, Fig.~\ref{fig:HyperParameters} (a) shows that 0.025 yields the best overall metrics. Conversely, excessively high (e.g., 0.045) or low (e.g., 0.005) weights lead to worse D\textsubscript{obj} and Col\textsubscript{obj} values.
Fig.~\ref{fig:HyperParameters} (b) demonstrates that both D\textsubscript{obj} and Col\textsubscript{obj} are optimal when the weight of O2O losses is approximately 0.05.
Fig.~\ref{fig:HyperParameters} (c) shows that the optimal results are achieved when the weight of the C2C loss is approximately 0.04.

\subsection{Use Cases and Applications}\label{sec:app}
Fig.~\ref{fig:application} presents an end-to-end workflow that takes as input a floorplan (a) and a set of furniture objects to generate an indoor layout (b). Subsequently, additional specified objects are integrated into the initial scene to produce a coherent interior arrangement (c). To satisfy user-defined spacing requirements, the system incorporates spatial relationships and utilizes WDT nodes to automatically adjust furniture positions (d), thereby yielding a finalised layout that adheres to design specifications (e). The resulting output is a customised design solution that both respects user input and complies with predefined spatial relationships. Accordingly, this framework enables spatial relationship-aware layout generation while also supporting flexible, user-in-the-loop refinement.


\subsection{Limitations}\label{sec:limit}

\begin{figure}[htbp]
    \centering
    \includegraphics[width=0.88\linewidth]{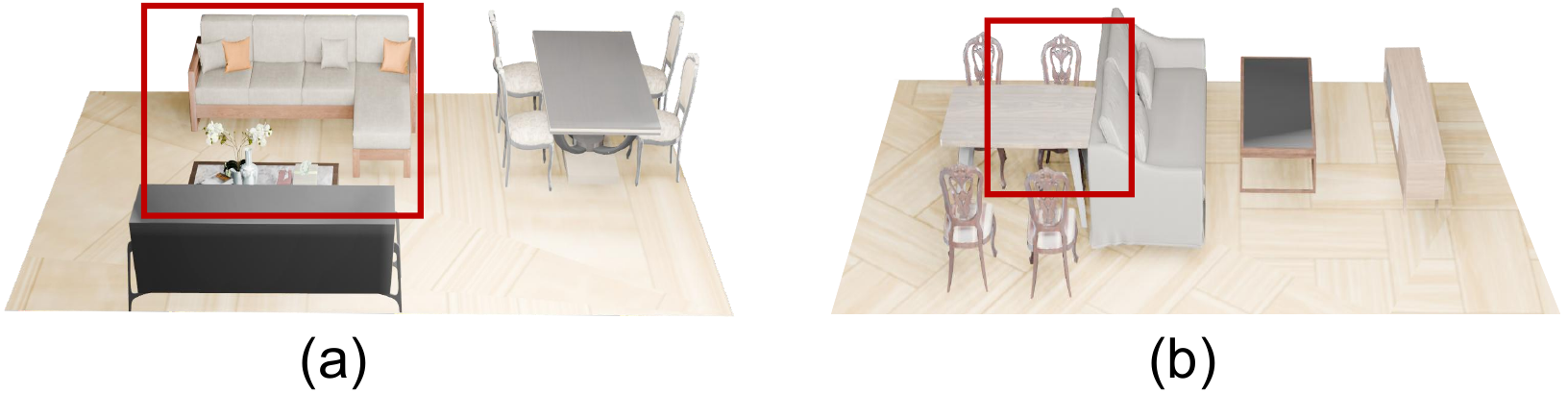}
        \caption{Cases of failure. We present representative failure cases of our scene generation results. The erroneous regions are highlighted with \textcolor{red}{red} boxes.}
    \label{fig:failure}  
\end{figure}

While \method{} demonstrates promising results in generating indoor layouts with better spatial relationships, there are still several limitations that need to be addressed. 
The method employs 3D bounding boxes to abstract away the geometric details of objects. However, real-world furniture often exhibits irregular shapes—for example, the sofa illustrated in Fig.~\ref{fig:failure} (a) is positioned far from the coffee table. 
The lack of explicit constraints between functional regions may result in generated objects intersecting with one another, as demonstrated in Fig.~\ref{fig:failure} (b). 
An additional core challenge is the difficulty of learning 3D spatial relationships with bounding box-based object abstractions, which leads to unnecessary ambiguity and inaccuracies. For instance, a table and a chair can be well-separated in the 3D space but usually have overlaps when using 3D bounding box-based representations. 
Therefore, it is interesting to explore more fine-grained 3D abstractions that can better reflect 3D layouts without the bother of too many shape details. 
Additionally, existing indoor datasets are significantly simpler than real-world scenes, both in terms of object diversity and layout complexity. 
Constructing large-scale layouts that include a wide variety of scenes, from residential to commercial spaces, will facilitate more realistic layout modeling. 
\section{Conclusion}
In this paper, we introduced \method{}, a novel framework for furniture layout generation. We leverage the indoor hierarchy to automatically extract object-to-region, object-to-object and corner-to-corner relationships and integrate the learning of these relationships into a generative diffusion framework. Our method aims to ensure realistic and functional layouts that respect practical relationships. Experimental results demonstrate that \method{} outperforms existing methods in a wide variety of metrics. Future work will explore adapting the hierarchical relational paradigm to multi-room or mixed-use environments to broaden its applicability and enable comprehensive, data-driven solutions for complex interior designs.

\bibliographystyle{IEEEtran}
\bibliography{main}
\end{document}